%% file: main.tex
\documentclass[twocolumn, switch]{article} 

\usepackage{preprint}

\usepackage{amsmath, amsthm, amssymb, amsfonts}

\usepackage[numbers,square]{natbib}

\usepackage[utf8]{inputenc}	
\usepackage[T1]{fontenc}	
\usepackage{xcolor}		
\usepackage[colorlinks = true,
            linkcolor = purple,
            urlcolor  = blue,
            citecolor = cyan,
            anchorcolor = black]{hyperref}	
\usepackage{booktabs} 		
\usepackage{nicefrac}		
\usepackage{microtype}		
\usepackage{lineno}		
\usepackage{float}			

\usepackage{lipsum}		

\usepackage{newfloat}
\DeclareFloatingEnvironment[name={Supplementary Figure}]{suppfigure}
\usepackage{sidecap}
\sidecaptionvpos{figure}{c}

\usepackage{titlesec}
\titlespacing\section{0pt}{12pt plus 3pt minus 3pt}{1pt plus 1pt minus 1pt}
\titlespacing\subsection{0pt}{10pt plus 3pt minus 3pt}{1pt plus 1pt minus 1pt}
\titlespacing\subsubsection{0pt}{8pt plus 3pt minus 3pt}{1pt plus 1pt minus 1pt}

\usepackage{tikz,xcolor,hyperref}

\usepackage{multirow}

\title{NODE IK: Solving Inverse Kinematics with Neural Ordinary Differential Equations \\ for Path Planning}

\input{commands}

\usepackage{authblk}

\author[1\thanks{\tt{psh117@snu.ac.kr}}]{Suhan Park}
\author[2\thanks{\tt{cadop@njit.edu}}]{Mathew Schwartz}
\author[1\thanks{\tt{park73@snu.ac.kr}}]{Jaeheung Park}

\affil[1]{Department of Intelligence and Information, Seoul National University, \\
Seoul, 08826, Korea}
\affil[2]{College of Architecture and Design, NJIT, \\
New Jersey, USA}

\begin{document}

\twocolumn[ 
  \begin{@twocolumnfalse} 
  
\maketitle

\begin{abstract}
        This paper proposes a novel inverse kinematics (IK) solver of articulated robotic systems for path planning. IK is a traditional but essential problem for robot manipulation. Recently, data-driven methods have been proposed to quickly solve the IK for path planning. These methods can handle a large amount of IK requests at once with the advantage of GPUs. However, the accuracy is still low, and the model requires considerable time for training. Therefore, we propose an IK solver that improves accuracy and memory efficiency by utilizing the continuous hidden dynamics of Neural ODE. The performance is compared using multiple robots.
\end{abstract}


\vspace{0.35cm}

  \end{@twocolumnfalse} 
] 


\section{Introduction}

An \textit{Inverse kinematics} (IK) problem is to find the joint space solution that satisfies the given Cartesian pose constraint. 
This process is important as most real-world tasks, such as furniture assembly \cite{park2022robotic} and roboscuplt \cite{schwartz2013robosculpt}, have Cartesian targets and provide a meaningful method of determining a necessary joint configuration to achieve the desired end-goal. By extending a single IK solution into a series of multiple IK solutions, one can construct a method for moving along arbitrary locations in Cartesian space -- creating a path of motion for a robot. 
For this reason, IK has been a topic of robotics research for decades, and the improved computational speed, configuration options, and accuracy have been the subject of numerous open-source initiatives. These include \eg an analytic inverse kinematics solver~\cite{diankov2010automated} for six degree-of-freedom (DoF) robots and a solver using numerical optimization~\cite{beeson2015tracik}. Although the analytical IK solution is fast, it can only be applied to a limited number of robots. In particular, the complexity of the IK solver drastically increases when redundant degrees of freedom are present, as weighting and other priorities must be implemented to reduce the solution space. On the other hand, numerical optimization methods can be used in all robots, but suffer from computational complexity. 
Therefore, there is a significant time overhead for solving multiple IK requests, such as the IK for a Cartesian space path. 

Finding joint configuration for the given Cartesian space path is one of the constrained motion planning problems \cite{kingston2019exploring}, which is known as a complex problem because it has an implicit constraint on the path. The end-effector pose constraint is nonlinear and implicit for most robots. Therefore, searching path considering this constraint costs considerable time. Prior studies have shown that the constrained motion planning problem can be solved faster using IK \cite{ikpath} \cite{closedchain}. Therefore, batch IK predictions can enhance searching path in the end-effector pose constrained space and provide various paths that can be used as a warm start (initial) path. 

\begin{figure}
    \centering
    \includegraphics[width=\linewidth]{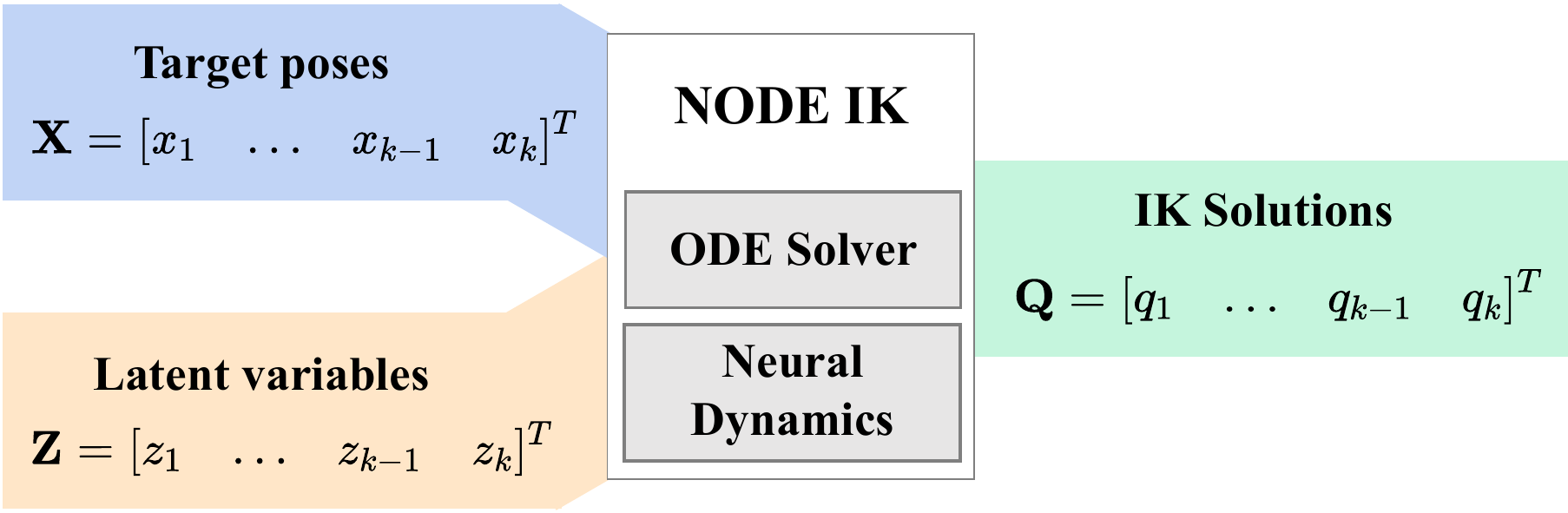}
    \caption{Objective of the proposed method: $k$-IK requests are solved at once, batch-IK. Utilizing GPU and neural ODE structure, multiple accurate  IK solutions can be quickly acquired from given target poses, such as the end-effector path. Latent variables are used to describe redundancy.}
    \label{fig:nodeik_def}
\end{figure}

Recently, data-driven methods~\cite{d2001learningik,rolf2010goallearningik} have been proposed to quickly find the multiple solutions required for path planning. By utilizing generative models, such as conditional variational autoencoder (CVAE) and conditional generative adversarial network (CGAN), a prior work \cite{approxmanifold} improved the sample drawing for constrained motion planning. However, CVAE and CGAN have known issues that manifold mismatch and mode collapse (CGAN), which eventually reduce available space and diversity. Kim and Perez \cite{kim2021naver} introduced an autoencoder style inverse kinematics network and density estimator using normalizing flows (NF). Similar to our work, IKFlow~\cite{ames2022ikflow} directly utilizes the conditional NF to improve the accuracy of the generated inverse kinematics. However, IKFlow needs extensive model parameters for highly accurate prediction because the model complexity depends on the layer width and depths. 
\begin{figure}
    \centering
    \includegraphics[width=0.83\linewidth]{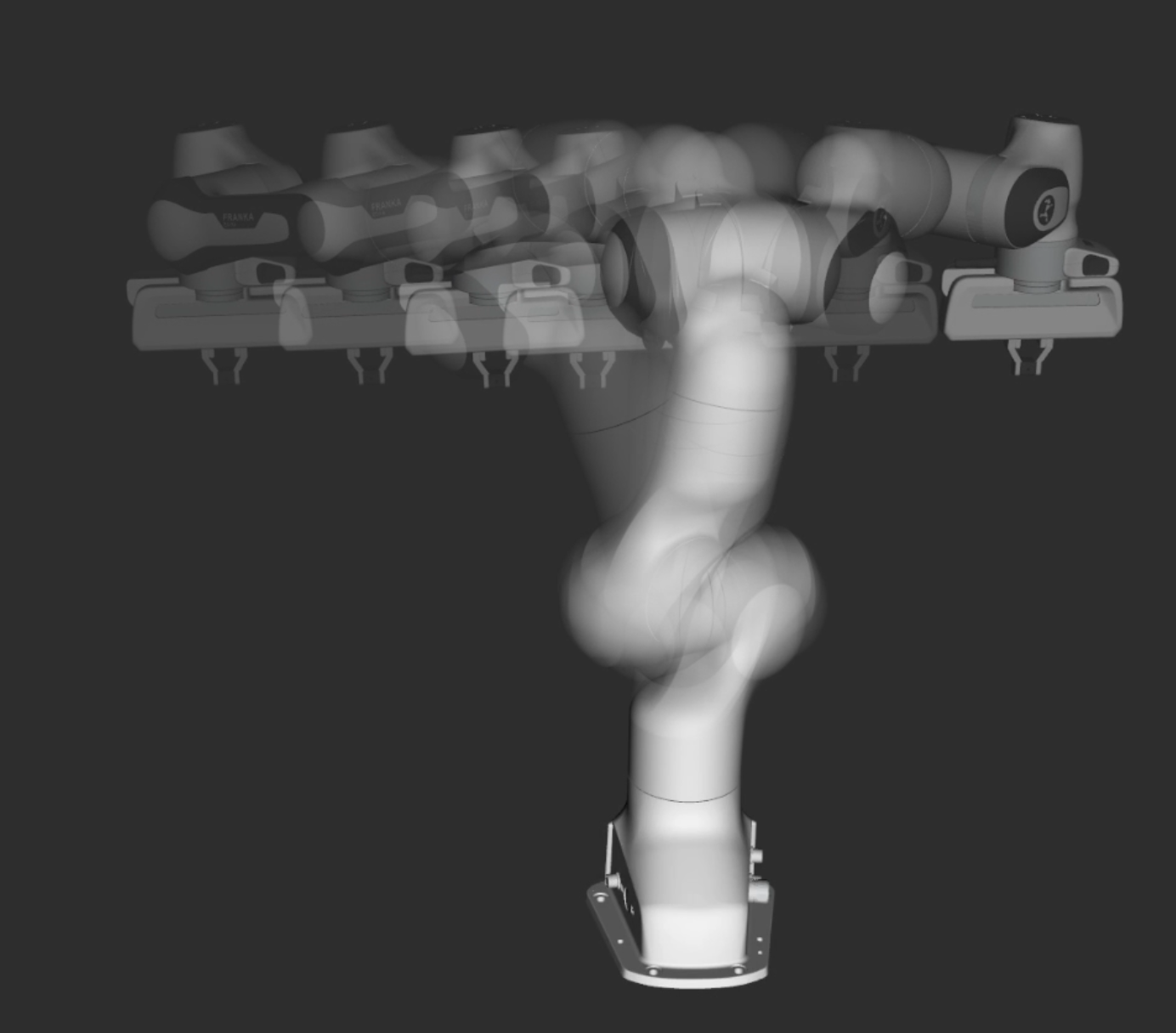}
    \caption{Cartesian pose constrained path from single batch IK prediction of NODE IK}
    \label{fig:node_ik}
\end{figure}
On the other hand, continuous layer models are known for their capacity to represent complex distribution with a relatively simple model. 
The continuous model can be structured using an ordinary differential equation (ODE) with hidden neural dynamics, which is called Neural ODE (NODE), and normalizing flows with NODE are called continuous normalizing flows (CNF) \cite{chen2018neural}. 
This paper proposes a novel IK method using the Neural ODE for lightweight and accurate inverse kinematics generation.
We refer to this process as NODE IK (Neural Ordinary Difference Equations for Inverse Kinematics). NODE IK is demonstrated on a redundant manipulator that has more than 6 DoF, as illustrated in Fig. \ref{fig:node_ik}. The objective of NODE IK is to find multiple IK solutions quickly from given target poses and latent variables that can be sampled in tractable normal distribution as shown in Fig. \ref{fig:nodeik_def}. In addition, the proposed model deals with multi-target IK as well as single-target IK using the expressive NODE model. As an application of the batch-IK, finding a joint path for the given Cartesian end-effector path is demonstrated to show the efficiency of the batch-IK.

The contribution of this paper is to reduce the number of model parameters while achieving better accuracy compared to the state-of-the-art method. In addition, multi-target IK has been demonstrated whose joint configuration comprises multi-branch kinematic chains, not single chains. For example, a joint configuration for dual arm tasks of humanoid robots includes common joints, such as waist joints, connecting the pelvis frame to each end-effector frame. Finally, as the objective of the proposed method is a large number of IK resolutions, we demonstrate the path generation results of the Cartesian pose constrained planning problem using the proposed method. Technically, integrating cutting-edge open source software, NODE IK is provided with GPU acceleration for both model training and forward kinematics. This enables online dataset generation in the training process. 

\section{Background}

\subsection{Inverse Kinematics}
Inverse kinematics is, as the name implies, the inverse of forward kinematics defined by a function $f : \R^n \rightarrow SE(3)$, mapping joint configurations $q \in \R^n$ to an end-effector pose $x \in SE(3)$, where $n$ is the number of joints. 
The methods for the IK problem can be divided into analytic and numerical approaches. An analytic approach directly solves the joint configurations using geometric relationships. Unfortunately, a geometry analysis for inverse kinematics of a robot with more than 6 DoF is challenging and often not feasible. On the other hand, numerical methods can be trivially implemented and used for all single-chain robot configurations. The Newton-Rhapson method \cite{gupta1985improved} and sequential quadratic programming \cite{beeson2015tracik} are well-known algorithms for solving the inverse kinematics problem. However, these methods take a considerable time for a large number of solutions -- a requirement for a smooth joint path through a Cartesian target path.

There can be infinite solutions for a single target pose for a redundant robot (one in which a typical arm has more than six DoF). Thus, introducing a latent variable $z \in \R^n$, the unique solution to the inverse kinematics can be determined as $q = f^{-1}(x,z)$. Note that the latent variable has the same dimension as the configuration because NF require the same dimension for the invertibility of the function. In this paper and prior work \cite{ames2022ikflow}, NF is utilized to transform between latent variable $z$ and joint configuration $q$.

\begin{figure*}
    \centering
    \includegraphics[width=\linewidth]{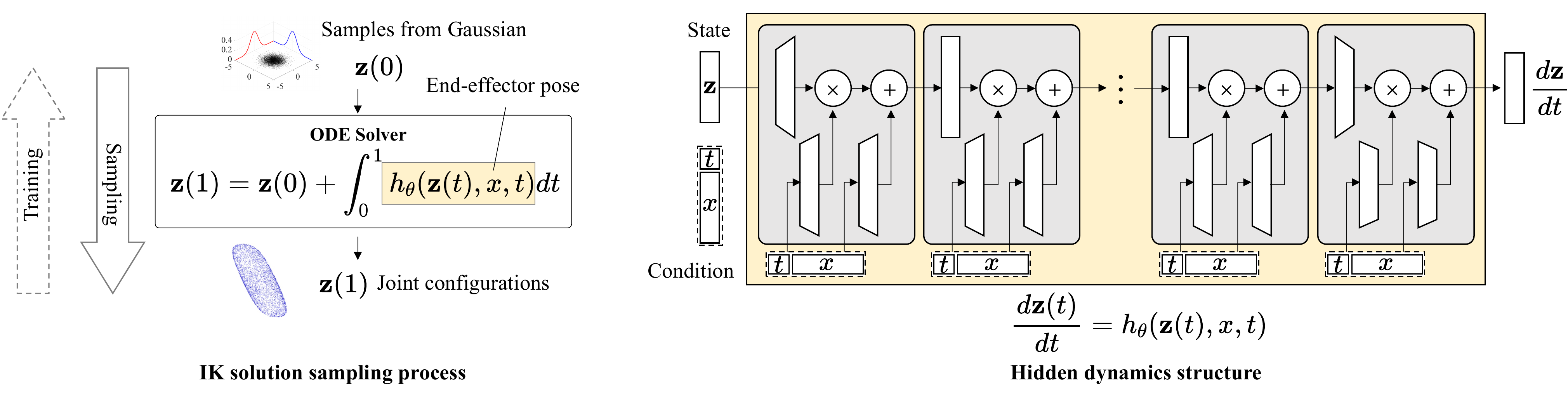}
    \caption{Structure of Node IK. Left figure is sampling process in the complex joint configuration distribution (IK). Right figure is structure of hidden dynamics.}
    \label{fig:architecture}
\end{figure*}
\subsection{Normalizing Flows} \label{sec:nf}
The normalizing flows (NF) \cite{rezende2015normalizingflows}, which are used in prior work \cite{ames2022ikflow}, are generative models with a tractable distribution for efficient density estimation and sampling in the complex distribution.
Using a tractable probability density function, such as normal distribution, NF are used to estimate the complex target distribution. When a sample is drawn from a tractable distribution, for example, $ \xp_t \sim \Nor(0,I)$, a model with learnable parameters $f_\theta : \R^D \rightarrow \R^D$ transforms the sample $\xp_t$ to a desired state $\xp_c$ in the complex distribution:

\begin{equation}
    \xp_c = f_\theta (\xp_t). \label{eq:transform_func}
\end{equation}

Conversely, the training process proceeds through the inverse function of the model $g_\theta = f_\theta^{-1} $ , maximizing log-likelihood. 
Let $\xp_t, \xp_c$ be the random variables in tractable and complex distribution and $g_\theta:\R^D \rightarrow \R^D$ be an invertible function where $\xp_t = g_\theta(\xp_c)$. Then, $p(\xp_t)$ is known by tractable distribution, and the change of variable theorem enables the computation of exact changes in probability:
\begin{equation}
    p(\xp_c)=p(\xp_t) \left| \det{\frac {\partial f}{ \partial \xp_t}} \right|^{-1}.
\end{equation}
And the log probability is
\begin{equation}
    \log{p(\xp_c)}=\log{p(\xp_t)} - \log \left| \det{\frac {\partial f}{ \partial \xp_t}} \right|.
\end{equation}
The model parameter $\theta$ is updated to maximize $\log{p(\xp_c)}$.

The prior work IKFlow \cite{ames2022ikflow} modeled the transform functions by a composition of the invertible functions:
\begin{equation}
    f_\theta= f_1 \circ \dots \circ f_{N-1} \circ f_N.
\end{equation}
The model depth $N$ needs to be large enough for an expressive model. However, the number of parameters increases when the model's depth $N$ increases.

\subsection{Neural ODE and Continuous Normalizing Flows}

A neural network for hidden dynamics $h_\theta$ representing ODE: 

\begin{equation}
    \frac{d\z (t)}{dt}=h_\theta(\z (t),t),
    \label{eq:node}
\end{equation}
which is called Neural ODE \cite{chen2018neural}, is known for its ability that estimates complex distribution using a relatively simple model.
Starting from the initial value, which is used as an input layer, $\z(t_0)$, the desired value $\z(t_1)$ is the solution to the initial value problem at some time $t_1$: 
\begin{equation}
    {\z(t_1)}={\z (t_0)} + \int_{t_0}^{t_1}{h_\theta \left(\z (t), t \right) dt}. \label{eq:cnf_transform}
\end{equation}
A black-box differential equation solver can compute the desired value, automatically evaluating the hidden dynamics $h_\theta$ as needed with the desired accuracy. This structure lets the hidden dynamics $h_\theta$ form continuous vector fields for transforming from the initial state to the desired state. The gradients for training are computed using the adjoint sensitivity method. Let $\Loss({\z(t_1)})$ be a loss function for the output of ODE solver: 
\begin{equation}
    \Loss({\z(t_1)})=\Loss\left({\z (t_0)} + \int_{t_0}^{t_1}{h_\theta \left(\z (t), t \right) dt}\right). \label{eq:loss}
\end{equation}
Utilizing the adjoint ${\mathbf{a}}(t)=\partial \Loss / \z(t)$, the gradient with regard to $\theta$ for optimizing the hidden neural dynamics:
\begin{equation}
    \frac{d\Loss}{d\theta}=-\int_{t_1}^{t_0}{\mathbf{a}}(t)^T \frac{\partial h_\theta \left(\z (t), t \right)}{\partial \theta}~dt. \label{eq:dtheta}
\end{equation}

This neural ODE form can be used for NF structures, which is called continuous normalizing flows (CNF) \cite{chen2018neural}.
Let $\z(t_0), \z(t_1)$ be the random variables in tractable and complex distribution as in Sec. \ref{sec:nf}~~. Then (\ref{eq:cnf_transform}) can be used as a transform function like (\ref{eq:transform_func}). 
This form is easily invertible by reverting $t$. Thus, $h_\theta$ does not need to be bijective for NF. In addition, this structure can be seen as having infinitely deep recurrent layers, enabling the expressive model. Note that time $t$ does not indicate robot trajectory time but the degree of transformation from $z$ to $q$. We will describe $\z(t=0)$ as $z$ and $\z(t=1)$ as $q$ in the rest of the paper.

CNF do not require determinant computation of the Jacobian because the change in log-density follows the instantaneous change of variables formula \cite{chen2018neural}:

\begin{equation}
    \frac{\partial \log {p(\z (t))}}{\partial t} = - {\mathrm{Tr}} \left( \frac {\partial h_\theta}{ \partial \z (t)}\right). \label{eq:tracejaco}
\end{equation}
Therefore, the target log-density is
\begin{equation}
    \log{p(\z(t_1))}=\log{p(\z (t_0))} - \int_{t_0}^{t_1}{{\mathrm{Tr}}\left(\frac{\partial h_\theta}{\partial \z (t)}\right)dt}, \label{eq:logdensity}
\end{equation}
which is solved by ODE solver. To maximize the log-likelihood, the loss function is
\begin{equation}
    {\mathcal{L}} = -\log{p(\z (t_1))}. \label{eq:loss}
\end{equation}

Additionally, free-form Jacobian of reversible dynamics (FFJORD) \cite{grathwohl2018ffjord} was proposed to reduce computation costs of the Jacobian trace of (\ref{eq:tracejaco}). This method utilizes Hutchinson's trace estimator \cite{hutchinson1989stochastic}, which allows a scalable unbiased estimation of the log-density. While computing Tr costs $\timeO(n^2)$, FFJORD can approximate the log-density at cost $\timeO(n)$.

\section{Methods}
\subsection{Overview}
The proposed model exploits the aforementioned property of the NF and Neural ODE for lightweight and accurate IK prediction. The proposed method uses conditional CNF to learn the configuration distribution for a given Cartesian pose $x$.
The proposed method is implemented on top of SoftFlow \cite{kim2020softflow} architecture for conditioning and FFJORD \cite{grathwohl2018ffjord} for fast training. NVIDIA Warp \cite{warp2022} is utilized to compute the forward kinematics using GPU resources. 

\subsection{Architecture}
The proposed model utilizes conditional CNF using neural ODE:
\begin{equation}
    \frac{d \z (t)}{dt}=h_\theta(\z(t),x,t).
    \label{eq:node}
\end{equation}
The hidden dynamics model are conditioned by $t$ and $x$ for the dynamics corresponding to the end-effector pose. 
During a training procedure, instantly generated dataset is utilized.  

Joint configurations $q$ are sampled in uniform distribution within the joint range. And, the forward kinematics for the joints is computed online: 
\begin{align}
    q &\sim U(\underline{q},\bar{q}), \\
    x  &= f(q),
\end{align}
where $\underline{q}$ and $\bar{q}$ are the lower and upper bounds of the joints.
Then, the trace of Jacobian (\ref{eq:tracejaco}) and the target value $\z(0)$ of the current network model is computed as following relationship: 
\begin{align}
    \z(1) &= q, \\
    z \triangleq \z(0) &= \z(1) - \int_{0}^{1}{h_\theta(\z(t),x,t) dt}.
\end{align}

The proposed model uses a normal distribution as a tractable base distribution: $ z \sim {\mathcal{N}}(0,I)$ to easily sample and calculate the log density $\log{p(\z(0))}$. The model is directly updated by (\ref{eq:dtheta}) with the loss function (\ref{eq:loss}).

The structure of neural dynamics model $h_\theta$ is shown in Fig. \ref{fig:architecture}. Because CNF do not require Jacobian determinant, the structure of hidden dynamics can be simply designed. Every hidden layer is conditioned in affine form. 

\begin{figure}
    \centering
    \includegraphics[width=\linewidth]{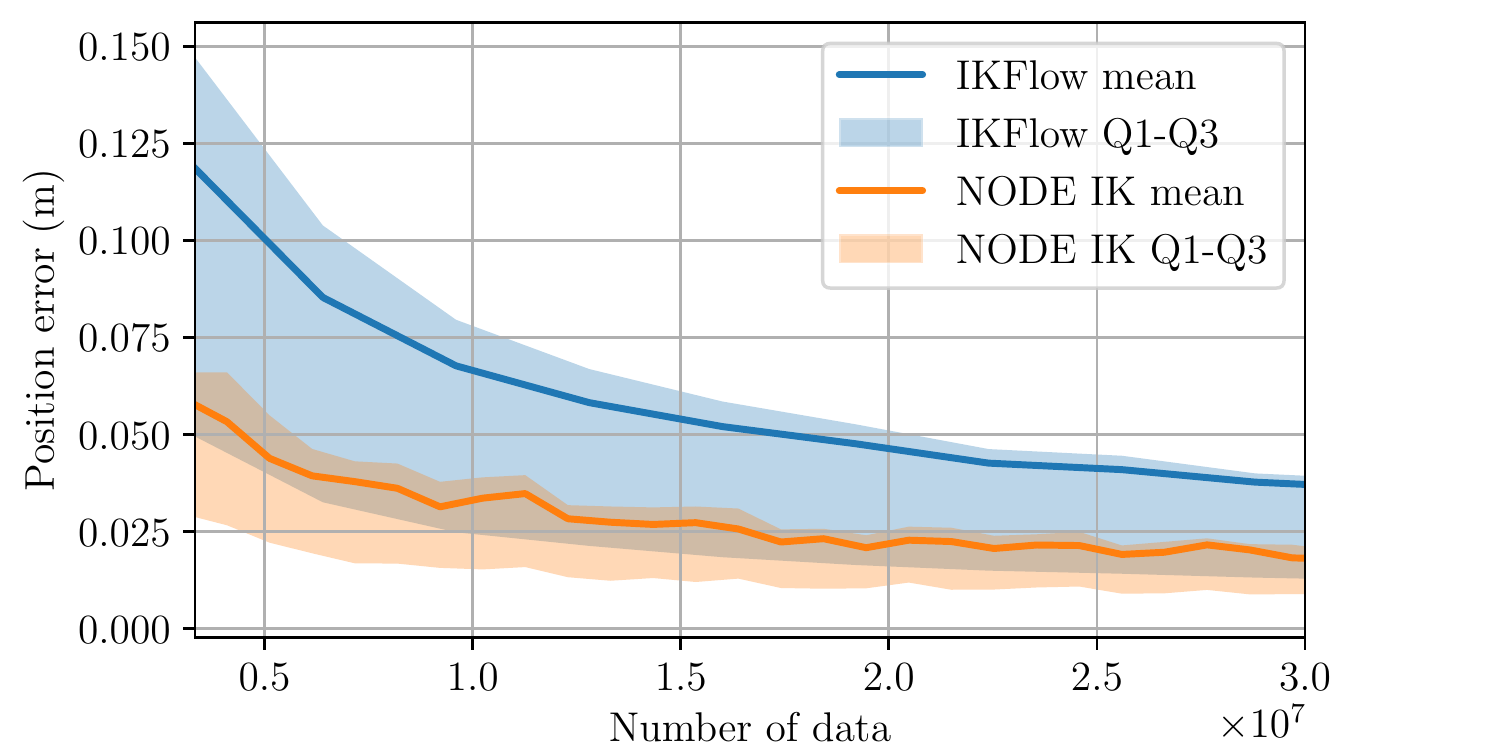}
    \caption{Position error while training}
    \label{fig:position_error}
\end{figure}
\subsection{Batch IK and Path generation}
When acquiring the batch IK for $k$ targets, a set of latent variable ${\mathbf{Z}} = [z_1, \dots, z_{k-1}, z_{k}]^T \in \R^{n \times k}$ are randomly sampled in normal distribution and target poses ${\mathbf{X}}=[x_1, \dots, x_{k-1}, x_k]^T \in SE(3)^{km}$ are set for the given path, where $m$ is the number of IK targets, \eg dual end-effector targets for humanoid robots. The output of the NODE IK is the corresponding joint configurations $ {\mathbf{Q}}=[q_1, \dots, q_{k-1}, q_{k}]^T \in \R^{n \times k}$. Fig. \ref{fig:nodeik_def} illustrates this relationship. This process is a one-time calculation taking advantage of GPUs.

When generating IK solutions for the given Cartesian path, $z$ should change continuously. The same or continuous change of $z$ is required for the continuous joint solution path. However, because the NF can learn multi-modal distribution, the path can be discontinuous even with the continuous change of the input variables. Therefore, after batch generation, every path is checked whether the path is continuous. 

\section{Results}

\subsection{Performance}

The proposed method and IKFlow were compared in accuracy, prediction time, and the number of model parameters. 

The accuracy was measured using 7 DoF Franka Emika Panda robot. Model hyperparameters for IKFlow \cite{ames2022ikflow} were set to 1024 wide network width, 12 coupling layers as they claimed the best performance with these parameters with the robot. 
The hidden dynamics of NODE IK comprised four 1024 width layers.
The accuracy was evaluated from 1000 sampled Cartesian target poses $x$ and the predicted 250 joint solutions for each target pose as in the experiments of previous work \cite{ames2022ikflow}. By passing the joint solutions to the forward kinematics function, the end-effector pose of predicted joint solutions $\hat{x}$ was computed.

\begin{figure}
    \centering
    \includegraphics[width=\linewidth]{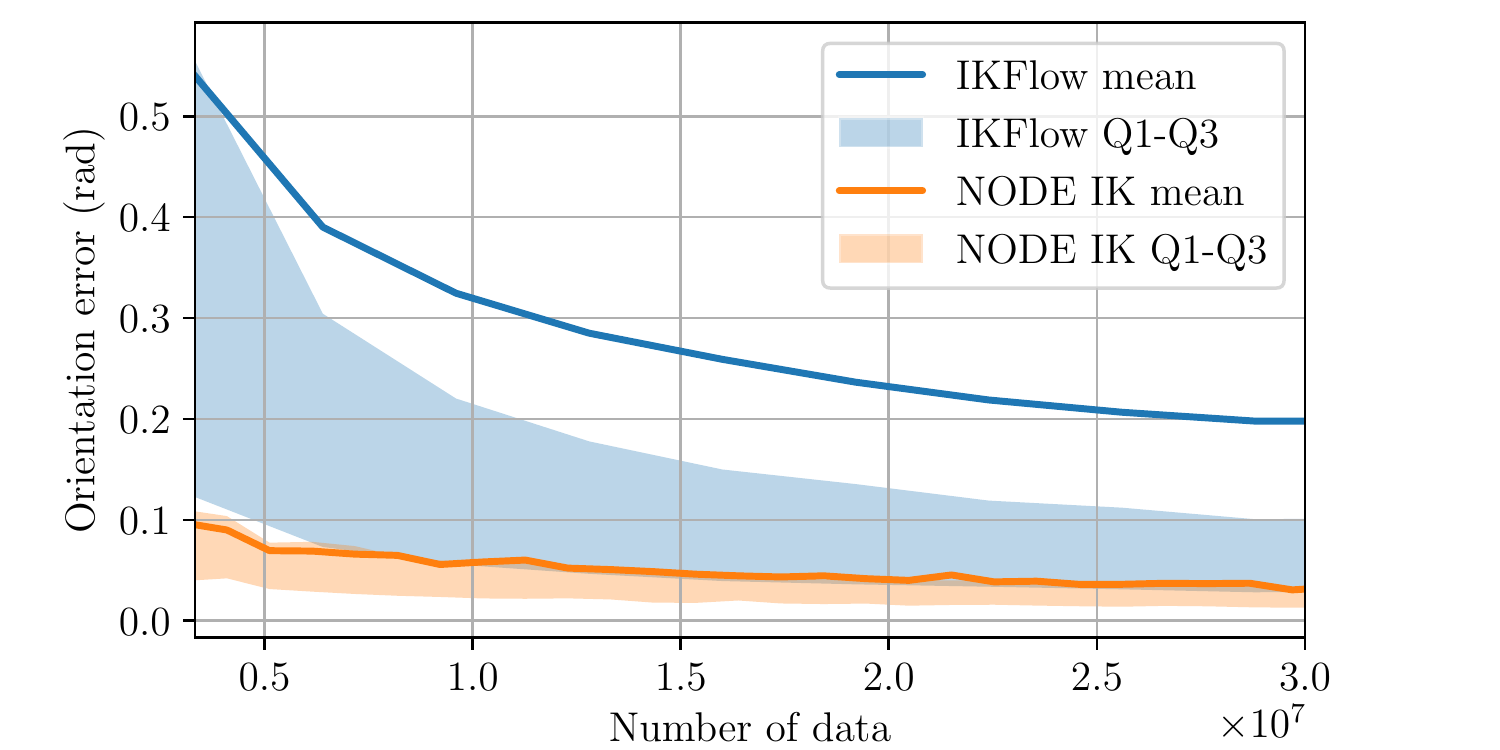}
    \caption{Orientation error while training}
    \label{fig:orientation_error}
\end{figure}

NODE IK was able to learn faster and more accurately than IKFlow with the same number of data passed to the network during the training process. Fig. \ref{fig:position_error} illustrates the position error of the end-effector poses $||x_p-\hat{x}_p||_2$. The position error of NODE IK was approximately half compared to that of IKFlow. Fig. \ref{fig:orientation_error} shows the orientation errors, which is quaternion geodesic distance between two poses. The orientation error of NODE IK was much lower than IKFlow. Outlier data with a large error existed among the generated solutions by IKFlow, which adversely affected the average value. 

Although NODE IK had high expressive power, it required a few parameters. The number of model parameters of NODE IK was 3,316,320 for achieving the sub-centimeter error, while that of IKFlow was 50,934,364. NODE IK had 6.5 \% parameters compared to IKFlow. 

However, NODE IK took more time for prediction than IKFlow because multiple prediction steps of the hidden dynamics are necessary for the ODE solver. System setup for the evaluation was with Intel i7-10700K CPU and NVIDIA RTX 2080 Ti GPU. NODE IK and IKFlow took 85.93 ms and 14.97 ms, respectively, to find 1000 IK solutions. 
Still, the IK generation speed was faster than that of TRAC-IK \cite{beeson2015tracik}, which is a numerical IK solver. Numerical solvers, TRAC-IK (C++) and Klampt IK (Python) took 1 s and 2.6 s, respectively, for 1000 IK solutions. 

\begin{figure*}[!tb]
    \centering
    \includegraphics[width=1.0\linewidth]{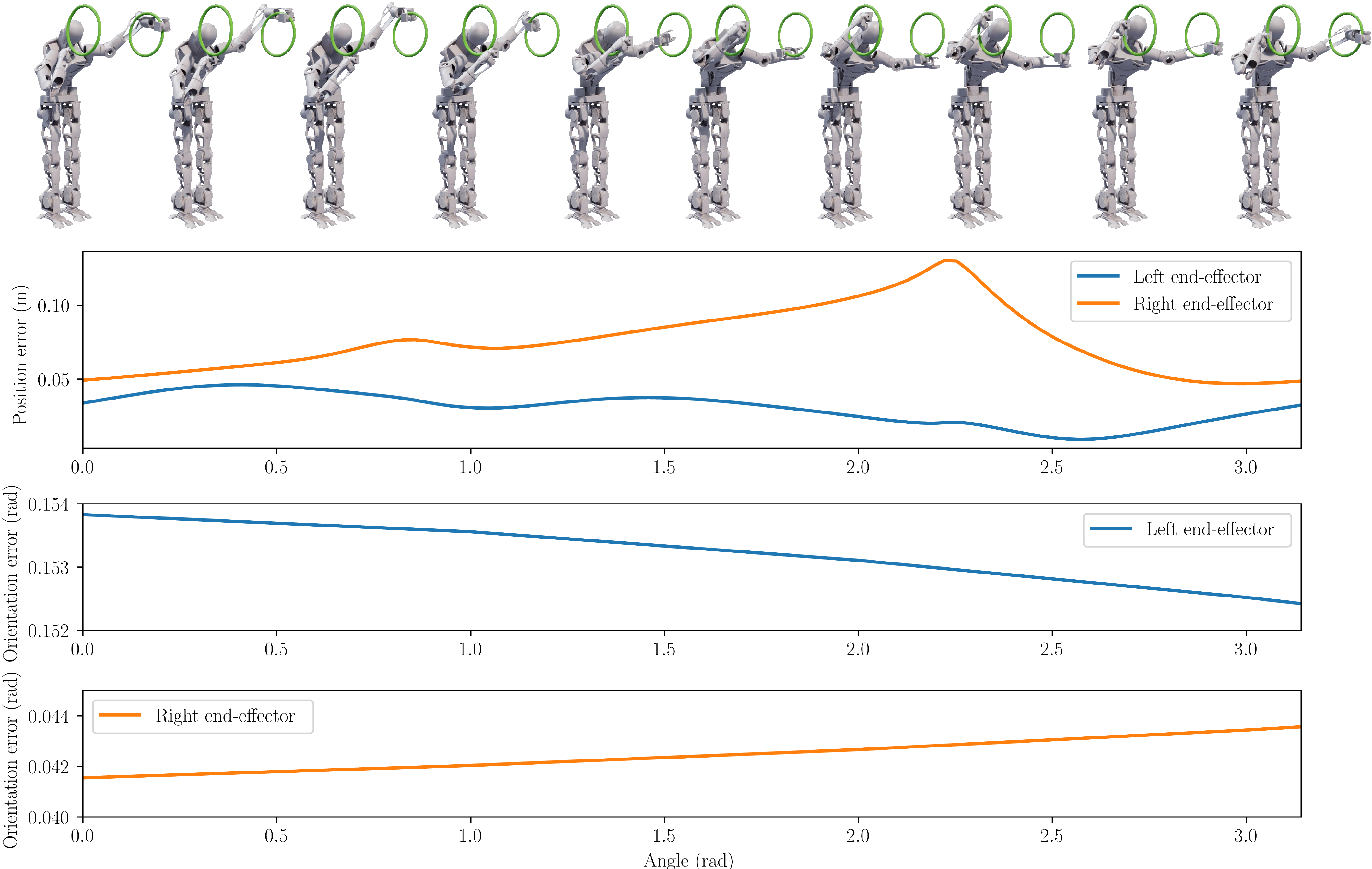}
    \caption{First row illustrates generated IK solutions following dual circle paths. Green circles are desired end-effector paths. Second row shows position error of the dual target path. Third and fourth rows show orientation error of the dual target path.}
    \label{fig:tocabi_circle}
\end{figure*}

\subsection{IK for path generation}
Path generation utilizing the advantages of batch IK was performed to evaluate the performance of the proposed model. Franka Emika Panda and TOCABI were used for the path generation. TOCABI is a 33 DoF humanoid robot, but, only upper body joints (arm and waist have eight and three joints, respectively) are used for IK. Target paths are given arbitrarily, which are circular paths. Orientation of the end-effector was fixed along the path. $z$ was sampled in $\Nor(0,I)$ once and all IK solutions were acquired using each target pose and the same $z$. The discontinuous paths were rejected. Path errors are the average of position and orientation errors at each target pose. Especially, NODE IK was able to learn dual target IK for humanoid robots. 19 joints are used for dual target IK, which includes three  shared waist joints. 

The average path errors are listed in Tab. \ref{tab:path_errors}~~~. Overall, NODE IK showed less error than IKFlow. As the IK complexity increased, the errors increased. A path generated from NODE IK is shown in Fig. \ref{fig:tocabi_circle} as a qualitative result. The position and orientation errors differed according to the targets, but the relative change was small. Especially, the orientation error was barely changed. NODE IK had better performance for learning orientation conditions than IKFlow.

\begin{table}
\caption{Average path errors. \textbf{F}: IKFlow, \textbf{N}: NODE IK. \textbf{S}: Single target, \textbf{D}: Dual target.}
\centering
\begin{tabular}{ cccccc }
\toprule
\multirow{2}{*}{Robots} & \multicolumn{2}{c}{Position (\textbf{mm})} & \multicolumn{2}{c}{Orientation (\textbf{deg})} \\
 & \textbf{F} & \textbf{N} & \textbf{F} & \textbf{N}  \\
\midrule
    Panda & 8.15 & 4.39 & 0.77 & 0.32\\ 
\hline
TOCABI (\textbf{S}) & 27.11 & 14.31 & 6.64 & 0.82 \\ 
\hline
TOCABI (\textbf{D}) & - & 48.09 & - & 8.71 \\
\hline
\toprule
\label{tab:path_errors}
\end{tabular}
\end{table}

\section{Conclusion}

The proposed method has 93\% fewer parameters than the state-of-the-art method IKFlow, allowing the highly accurate learning-based IK for the low memory system. In addition, from the increased complexity, NODE IK could demonstrate dual arm tasks. 
However, the prediction took longer time than the state-of-the-art method because the ODE solver requires multiple predictions of the hidden dynamics network. We expect that the performance of NODE IK can be improved with future work and contributions from the machine learning community by implementing optimized solvers. 

\bibliography{ref}
\bibliographystyle{plain}

%

\end{document}

%% file: commands.tex
\usepackage{xcolor}
\usepackage{xspace}

\newcommand{\eg}{\emph{e.g.},\xspace}

\newcommand{\R}{{\mathbb{R}}}
\newcommand{\Nor}{{\mathcal{N}}}
\newcommand{\Loss}{{\mathcal{L}}}
\newcommand{\timeO}{{\mathcal{O}}}
\newcommand{\z}{{\mathbf{z}}}
\newcommand{\xp}{{\mathbf{x}}}

%% file: main.bbl
\begin{thebibliography}{10}

\bibitem{approxmanifold}
Cihan Acar and Keng~Peng Tee.
\newblock Approximating constraint manifolds using generative models for
  sampling-based constrained motion planning.
\newblock In {\em 2021 IEEE International Conference on Robotics and Automation
  (ICRA)}, pages 8451--8457, 2021.

\bibitem{ames2022ikflow}
Barrett Ames and Jeremy Morgan.
\newblock {IKFlow}: Generating diverse inverse kinematics solutions.
\newblock {\em IEEE Robotics and Automation Letters}, 2022.

\bibitem{beeson2015tracik}
Patrick Beeson and Barrett Ames.
\newblock {TRAC-IK}: An open-source library for improved solving of generic
  inverse kinematics.
\newblock In {\em 2015 IEEE-RAS 15th International Conference on Humanoid
  Robots (Humanoids)}, pages 928--935. IEEE, 2015.

\bibitem{ikpath}
D.~Bertram, J.~Kuffner, R.~Dillmann, and T.~Asfour.
\newblock An integrated approach to inverse kinematics and path planning for
  redundant manipulators.
\newblock In {\em Proceedings 2006 IEEE International Conference on Robotics
  and Automation, 2006. ICRA 2006.}, pages 1874--1879, 2006.

\bibitem{chen2018neural}
Ricky~TQ Chen, Yulia Rubanova, Jesse Bettencourt, and David~K Duvenaud.
\newblock Neural ordinary differential equations.
\newblock {\em Advances in neural information processing systems}, 31, 2018.

\bibitem{diankov2010automated}
Rosen Diankov.
\newblock Automated construction of robotic manipulation programs.
\newblock 2010.

\bibitem{d2001learningik}
Aaron D'Souza, Sethu Vijayakumar, and Stefan Schaal.
\newblock Learning inverse kinematics.
\newblock In {\em Proceedings 2001 IEEE/RSJ International Conference on
  Intelligent Robots and Systems. Expanding the Societal Role of Robotics in
  the the Next Millennium (Cat. No. 01CH37180)}, volume~1, pages 298--303.
  IEEE, 2001.

\bibitem{grathwohl2018ffjord}
Will Grathwohl, Ricky~TQ Chen, Jesse Bettencourt, Ilya Sutskever, and David
  Duvenaud.
\newblock {FFJORD}: Free-form continuous dynamics for scalable reversible
  generative models.
\newblock In {\em International Conference on Learning Representations}, 2018.

\bibitem{gupta1985improved}
K~Gupta and Kazem Kazerounian.
\newblock Improved numerical solutions of inverse kinematics of robots.
\newblock In {\em Proceedings. 1985 IEEE International Conference on Robotics
  and Automation}, volume~2, pages 743--748. IEEE, 1985.

\bibitem{hutchinson1989stochastic}
Michael~F Hutchinson.
\newblock A stochastic estimator of the trace of the influence matrix for
  laplacian smoothing splines.
\newblock {\em Communications in Statistics-Simulation and Computation},
  18(3):1059--1076, 1989.

\bibitem{closedchain}
Keunwoo Jang, Jiyeong Baek, Suhan Park, and Jaeheung Park.
\newblock Motion planning for closed-chain constraints based on probabilistic
  roadmap with improved connectivity.
\newblock {\em IEEE/ASME Transactions on Mechatronics}, pages 1--9, 2022.

\bibitem{kim2020softflow}
Hyeongju Kim, Hyeonseung Lee, Woo~Hyun Kang, Joun~Yeop Lee, and Nam~Soo Kim.
\newblock Softflow: Probabilistic framework for normalizing flow on manifolds.
\newblock {\em Advances in Neural Information Processing Systems},
  33:16388--16397, 2020.

\bibitem{kim2021naver}
Seungsu Kim and Julien Perez.
\newblock Learning reachable manifold and inverse mapping for a redundant robot
  manipulator.
\newblock In {\em 2021 IEEE International Conference on Robotics and Automation
  (ICRA)}, pages 4731--4737. IEEE, 2021.

\bibitem{kingston2019exploring}
Zachary Kingston, Mark Moll, and Lydia~E Kavraki.
\newblock Exploring implicit spaces for constrained sampling-based planning.
\newblock {\em The International Journal of Robotics Research},
  38(10-11):1151--1178, 2019.

\bibitem{warp2022}
Miles Macklin.
\newblock Warp: A high-performance python framework for gpu simulation and
  graphics.
\newblock \url{https://github.com/nvidia/warp}, March 2022.
\newblock NVIDIA GPU Technology Conference (GTC).

\bibitem{park2022robotic}
Suhan Park, Haeseong Lee, Seungyeon Kim, Jiyeong Baek, Keunwoo Jang,
  Hyoung~Cheol Kim, Myeongsoo Kim, and Jaeheung Park.
\newblock Robotic furniture assembly: task abstraction, motion planning, and
  control.
\newblock {\em Intelligent Service Robotics}, pages 1--17, 2022.

\bibitem{rezende2015normalizingflows}
Danilo Rezende and Shakir Mohamed.
\newblock Variational inference with normalizing flows.
\newblock In {\em International conference on machine learning}, pages
  1530--1538. PMLR, 2015.

\bibitem{rolf2010goallearningik}
Matthias Rolf, Jochen~J Steil, and Michael Gienger.
\newblock Goal babbling permits direct learning of inverse kinematics.
\newblock {\em IEEE Transactions on Autonomous Mental Development},
  2(3):216--229, 2010.

\bibitem{schwartz2013robosculpt}
Mathew Schwartz and Jason Prasad.
\newblock Robosculpt.
\newblock In {\em Rob| Arch 2012}, pages 230--237. Springer, 2013.

\end{thebibliography}
